\documentclass[letterpaper, 10 pt, conference]{ieeeconf}  % Comment this line out if you need a4paper

\IEEEoverridecommandlockouts                              % This command is only needed if 
\usepackage{graphics} % for pdf, bitmapped graphics files
\usepackage{epsfig} % for postscript graphics files
\usepackage{mathptmx} % assumes new font selection scheme installed
\usepackage{times} % assumes new font selection scheme installed
\usepackage{amsmath} % assumes amsmath package installed
\usepackage{amssymb}  % assumes amsmath package installed
\usepackage{caption}
\usepackage{subcaption}
\usepackage{wrapfig}
\usepackage{soul}
\usepackage[dvipsnames]{xcolor}
\usepackage{tikz-cd}
\usepackage{cite}
\usepackage{tabu, booktabs, tabularx}

\definecolor{red}{RGB}{255, 0, 0}

\title{\LARGE \bf
Back to the Manifold: \\ Recovering from Out-of-Distribution States
}

\author{Alfredo Reichlin$^{1}$, Giovanni Luca Marchetti$^{1}$, Hang Yin$^{1}$, Ali Ghadirzadeh$^{2}$ and Danica Kragic$^{1}$% <-this % stops a space
\thanks{$^{1}$KTH Royal Institute of Technology,}%
\thanks{$^{2}$ Stanford University}%
}

\begin{document}

\maketitle
\thispagestyle{empty}
\pagestyle{empty}

%%%%%%%%%%%%%%%%%%%%%%%%%%%%%%%%%%%%%%%%%%%%%%%%%%%%%%%%%%%%%%%%%%%%%%%%%%%%%%%%
\begin{abstract}
Learning from previously collected datasets of expert data offers the promise of acquiring robotic policies without unsafe and costly online explorations.
However, a major challenge is a distributional shift between the states in the training dataset and the ones visited by the learned policy at the test time.
While prior works mainly studied the distribution shift caused by the policy during the offline training, the problem of recovering from out-of-distribution states at the deployment time is not very well studied yet.
We alleviate the distributional shift at the deployment time by introducing a recovery policy that brings the agent back to the training manifold whenever it steps out of the in-distribution states, e.g., due to an external perturbation. 
The recovery policy relies on an approximation of the training data density and a learned equivariant mapping that maps visual observations into a latent space in which translations correspond to the robot actions. We demonstrate the effectiveness of the proposed method through several manipulation experiments on a real robotic platform. Our results show that the recovery policy enables the agent to complete tasks while the behavioral cloning alone fails because of the distributional shift problem.

\end{abstract}

\section{Introduction}

Data-driven methods in robotics, including reinforcement learning (RL), are often challenged by expensive, slow and unsafe data collection on real systems~\cite{levine2018learning}. Offline solutions, such as \emph{Offline RL} \cite{levine2020offline} and \emph{Behavior Cloning} (BC) \cite{hussein2017imitation}, learn a control policy from a pre-collected dataset hence avoiding the problems of interacting with a physical robot. 
However, learning from a fixed offline dataset may compromise the capacity of the learner on dealing with novel situations not contained in the training dataset at the test time. 
Querying the trained policy on such out-of-distribution (OOD) inputs can exacerbate the compounding of the errors when the policy is subsequently applied to states evolved according to the previous state and action \cite{dagger}. Offline RL avoids this problem by constraining the learned policy to deviate minimally from the policy that collected the data \cite{zhou2020plas, kumar2020conservative, yu2021combo}. However, there is no mechanism for offline RL to recover from OOD conditions, for example, when starting at a random unseen initial state or being exposed to external perturbations during execution. 

One way to improve the performance of the agent in the deployment phase is to optimize an extra objective, e.g., by minimizing the uncertainty-level of the agent in predicting the next state \cite{ghadirzadeh2016sensorimotor, kahn2017uncertainty} which implicitly
helps the agent to stay in-distribution. However, designing such objectives is challenging and they may require learning probabilistic visual forward dynamic models that output a measure of uncertainty.

\begin{figure}[t]
\centering
\includegraphics[width=.75\columnwidth]{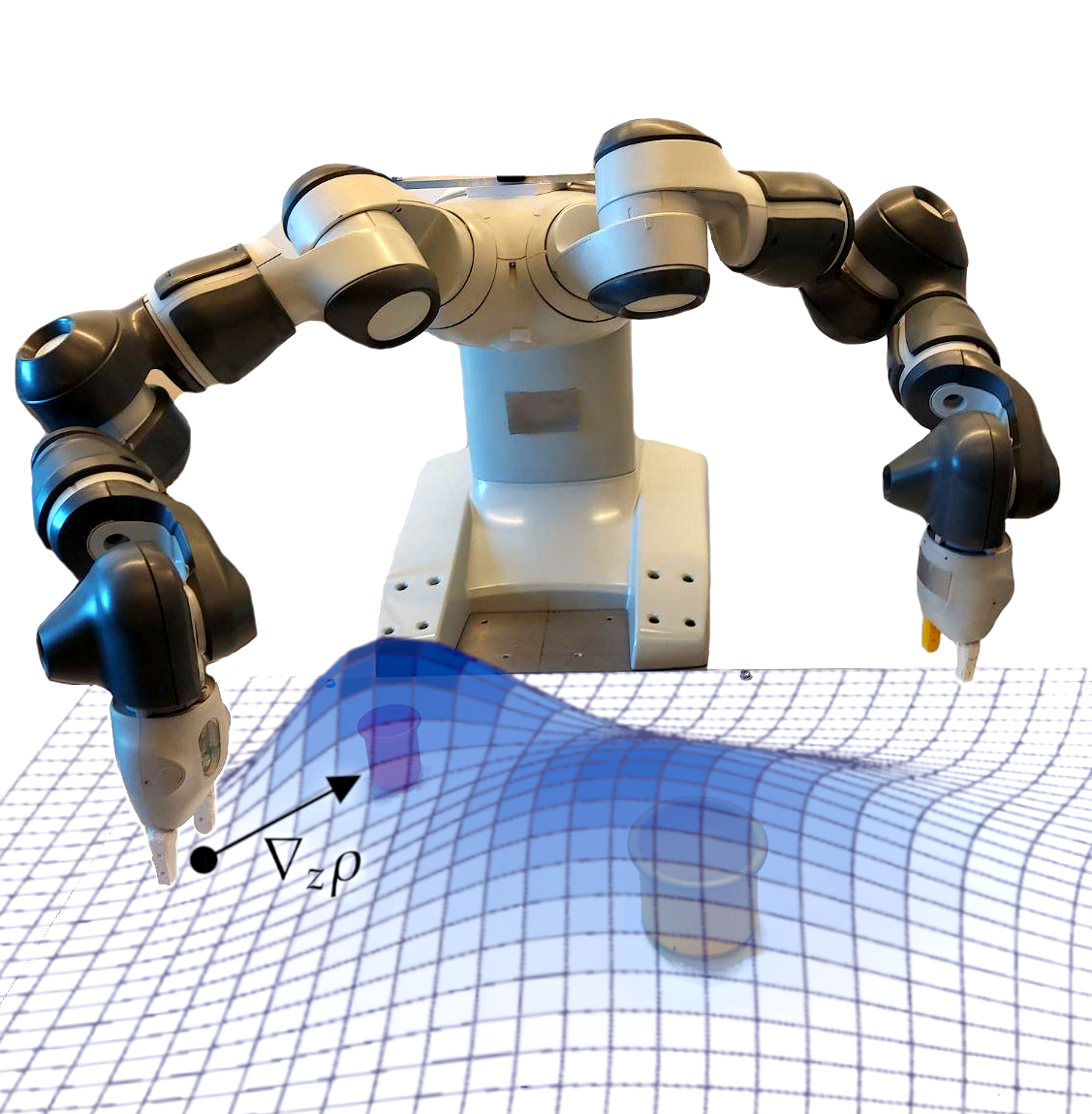}
    \caption{The proposed recovery policy performs gradient ascent on the estimated density $\rho$ of the demonstrations to get the agent back in-distribution.}\label{firstpagefigure}
\end{figure}

In this work, we propose a method to augment a policy trained by offline behavior cloning with a \emph{recovery policy} whose actions are computed by a gradient ascent on the estimated density of the training distribution. 
This is illustrated in Figure~\ref{firstpagefigure}, in which the robot is guided by the recovery policy providing action directions to stay in-distribution while approaching the task object. 
We achieve this by training a model that encodes input visual observations into a Euclidean latent space, where translations in this space correspond to robot actions, such as Cartesian displacements of the robot end-effector. The encoding has a property known as translational \emph{equivariance} that allows for the conversion of the aforementioned gradient of the estimated training data density into an action. Therefore, the recovery design benefits from a latent representation that (1) is low-dimensional, thus amenable to density estimation, and (2) is task-agnostic, i.e., it can be shared among other tasks.

We empirically demonstrate the feasibility of the proposed method on real-robotic visuomotor policy training tasks. Compared to a behavioral cloning policy, we show that the augmented policy improves the success rate on manipulation tasks. Additionally, when the robot is externally pushed OOD, it allows to recover and successfully complete the task. We also demonstrate how the trained latent equivariant representation can be shared among several tasks, making it task-agnostic. Our main contributions are: 
\begin{itemize}
    \item A method to augment policies trained with offline data to recover from OOD conditions through the gradient of a conditional density estimator.
    \item An empirical evaluation of the performances of the proposed method on real-robotic manipulation tasks.
\end{itemize}

\section{Related Work}\label{relwork}

\textbf{Offline policy learning} mainly fall into behavior cloning \cite{survey1} and offline RL \cite{levine2020offline}. The classic formulation of BC has a number of limitations when learning on real-world data. The most prominent of which, is called the compounding error \cite{pmlr-v9-ross10a} and occurs due to the sequential nature of the learning framework. This problem is even more profound when learning from small-sized datasets. There have been a number of works targeting this, however, they generally break the fully offline formulation \cite{dagger, abbeel2010inverse, ho2016generative}.

Offline RL, on the other hand, formulates the problem of learning a policy using offline data from an RL perspective. A naive application of RL on previously collected data results in either an high variance of the learning process \cite{precup2000eligibility} or wrong estimates of the expected return \cite{kumar2019stabilizing}. Possible solutions to this involve either constraining the learned policy to minimize the deviation from the one that collected the data (behavioral policy) \cite{fujimoto2019off, zhou2020plas, chen2022latent} or incentivizing the policy to avoid actions on the boundary of the training distribution by changing the reward function \cite{kumar2020conservative, yu2020mopo, yu2021combo}. Moreover, in case the agent happens to step OOD, or it is forcefully brought there, there is no explicit way to recover. Which is what we address in this work.

\textbf{Safety measures} in the context of a learned controller have been addressed in different forms \cite{brunke2021safe}. One common thread is the identification of \textit{safe} regions where the agent can operate and the use of a recovery policy. In \cite{lee2020guided}, unknown regions are defined by the uncertainty estimate of a perception module. When the agent steps there, a model-based reset policy is triggered. Differently from our method, they require a model of the objects and they assume the transition function is known in a subset of the state space. Other works instead assume to directly have access to a \textit{constraint function} from the environment or approximate it from data, which quantifies the safeness of states. A policy can then be learned to actively avoid such states \cite{thananjeyan2021recovery, srinivasan2020learning} or plan a trajectory that remains in the safe region of the state space \cite{thananjeyan2020safety, wilcox2022ls3, mitsioni2021safe}. Having access to the constraint function or data-points in unsafe states can, however, be problematic for robotic applications. A similar work to ours proposes to learn an approximation of the tangent space of the task manifold at any point \cite{li2018learning}. This can, in turn, be used to plan the overall trajectory. If some kind of perturbation occurs, the agent can project its current position in the manifold and plan a path to get back in. The projection is learned explicitly using a dataset of perturbed points. On the contrary, the way we learn the encoder allows us to automatically get the projection direction even from high-dimensional states like images.

\textbf{State Representation} for control has been widely studied in prior works \cite{lesort2018state, chen2020adversarial, hamalainen2019affordance}. Dividing the optimization of the representation from the policy has the advantage of easing the learning process and enables to constrain the policy formulation \cite{finn2016deep, ghadirzadeh2017deep}. Moreover, the learned representation can be re-used for different tasks assuming the underlying dynamics remain the same. In \cite{assael2015data} and \cite{ha2018world} an encoder is trained to compress the state representation while retaining all of the information. A transition model is then inferred on top of the representation to predict the dynamics of the environment. This formulation, however, produces a generic representation with no particular properties. To this end, more rigid constraints have been proposed. In \cite{goroshin2015learning} the representation is forced to evolve linearly in time, while in \cite{finn2016deep} the model outputs spatial features representing the observations. Moreover, by imposing a linear dynamic on the representation, the learned controller can be simplified and, under some assumptions, learned optimally \cite{watter2015embed, zhang2019solar}. Unlike our method, this dynamic cannot be directly converted into an action. \cite{jaques2021newtonianvae} train a variational autoencoder with the additional constraint of making the latent representation evolve according to Newtonian physics. This allows for classical controllers, like a PID, to be applied directly. In \cite{kipf2019contrastive}, they propose to learn an encoder and a transition model at the same time. Both models are learned such that the latent representation is equivariant to the transition model. Differently from all of these methods, we require our representation to be globally translational-equivariant in order to convert the gradient of the density estimator into a viable action.
\section{Background}

We study the problem of learning a policy using a Markov decision process (MDP). We assume the MDP to be fully observable and specified by the tuple $(S, A, T)$. Here $S$ is the set of observations representing the state of the environment, $A$ is the set of actions the agent can take, and $T: \ S \times A \rightarrow S$ is the transition function governing the change in observations when the agent performs an action. We can then formulate the problem of policy training as learning a map $\pi: \ S \rightarrow A$ by minimizing a cost function. Throughout this paper we define the set of states $S$ to be images and the set of actions $A \subseteq \mathbb{R}^n$ to be continuous translations of the agent's end-effector in the Euclidean space. A dataset $\mathcal{D}$ of experts' demonstrations is a collection of trajectories representing the robot completing a task in different conditions. This dataset can be considered as a collection of tuples $\mathcal{D} = \{(s, a, s')\}$ with $s'=T(s,a)$.

\subsection{Behavioral Cloning}

Behavioral Cloning is one of the most widely used imitation learning (IL) approaches due to its ease of implementation. It defines the cost function of the policy $\pi$ as a supervised learning loss. As such, the policy's parameters $\theta$ can be inferred by minimizing the Mean Squared Error (MSE) between its estimated actions and the ones in the dataset $\mathcal{D}$:
\begin{equation}
    \pi = \pi_{\theta^*}, \ \ \ \theta^* =\textnormal{argmin}_\theta \sum_{(s, a) \in \mathcal{D}} \| \pi_\theta (s) - a \|^2. 
\end{equation}
This simple IL strategy offers a number of advantages. First, it is easy to implement and stable to train. Second, it can be learned completely from offline data requiring neither access to the environment nor any knowledge of the transition model or a reward function.

\subsection{Equivariant Mapping}\label{equiv:backgroud}
In this section, we explain \emph{equivariance} as a property of a mapping $E$ that in our case maps an input observation $s$ into a latent representation $z \in Z$, i.e. $z = E(s)$.
Here, we consider two transition functions, $T: S \times A \rightarrow S$ and $T': \ Z \times A \rightarrow Z$.  
The map $E: \ S \rightarrow Z$ is defined to be equivariant with respect to the transitions $T$, $T'$ if \cite{cohen2016group, marchetti2022equivariant}:
\begin{equation}
    E(T(s, a)) = T'(E(s), a).
    \label{eq:equivariant_mapping}
\end{equation}

\begin{figure}[t]
\centering
\includegraphics[width=.5\columnwidth]{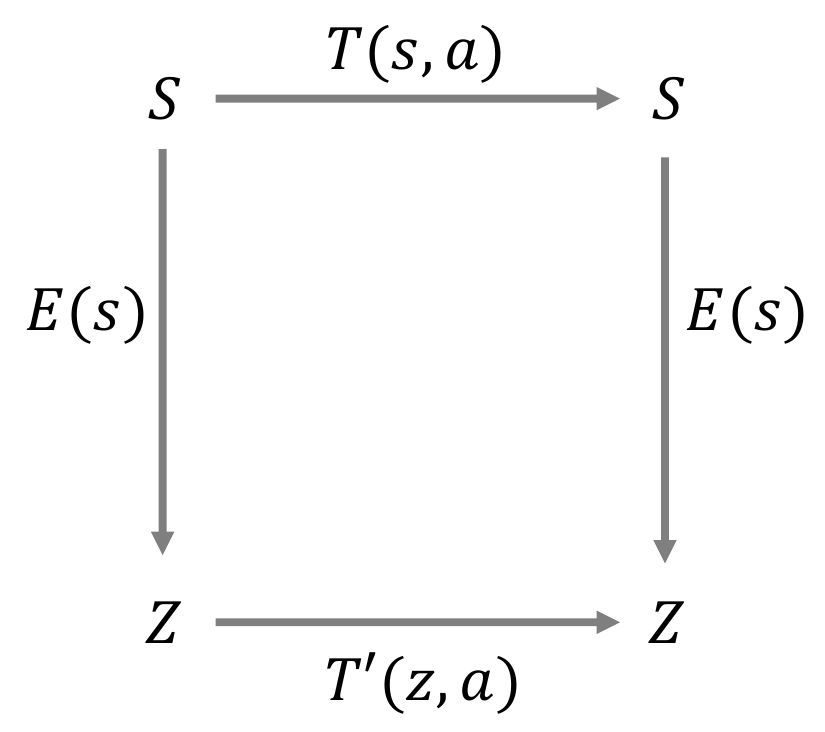}
    \caption{Commutative diagram illustrating equivariance as a property of the mapping $E$.}
    \label{fig:equivarence}
\end{figure}

Figure \ref{fig:equivarence} visually illustrates the equivariance property for the mapping $E$ and the transitions $T$ and $T'$. Intuitively, mapping the observation into the latent space followed by the transition $T'$ must result in the same latent value $z$ as applying the transition $T$ followed by mapping the resulting observation into the latent space.  

\begin{figure*}[tbh!]
    \centering
    \begin{subfigure}[b]{.45\linewidth}
        \centering
        \includegraphics[width=\linewidth]{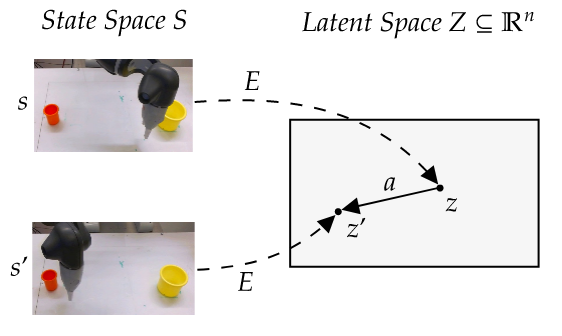}
    \end{subfigure}
     \begin{subfigure}[b]{.45\linewidth}
        \centering
        \includegraphics[width=\linewidth]{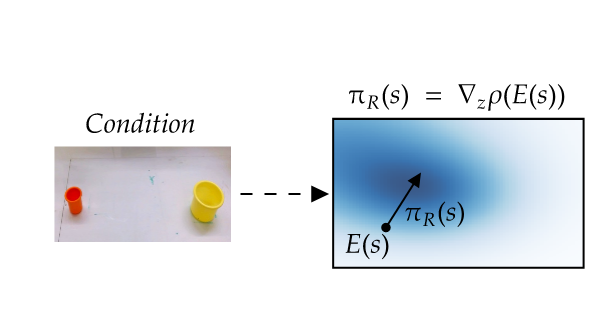}
    \end{subfigure}
    \caption{Overview of the proposed method. \textbf{Left}: the equivariant model $E$ maps the observation space (images) to the Euclidean latent space $Z$ contained in the action space $A$. Actions of the agent correspond to translations in $Z$. \textbf{Right}: the latent density $\rho$ is estimated conditioned on task-specific information. The recovery policy $\pi_R$ follows the gradient of $\rho$ and thus redirects the agent back to the training manifold. }
    \label{pipeline}
\end{figure*}

\section{Method}
In this section, we introduce our method that augments a policy trained with BC with a recovery strategy which can bring the agent back \emph{in-distribution}, where the BC policy performs optimally. We achieve this by introducing a \emph{recovery policy} $\pi_{R}: \ S \rightarrow A$ whose actions move the agent to the training manifold, i.e., within the support of the observations in the training dataset. 
The intuition is that the agent should follow the recovery policy to recover from OOD states, and the BC policy when in-distribution to complete the task.
Therefore, we propose to compute the actions given by each policy, and find the weighted sum of the actions to obtain the final policy output. 
The weights are computed according to a normalized estimation of the density of the training states. However, as we describe in section~\ref{densityest}, we estimate the density by first mapping the observation into a latent space $z = E(s)$, and then computing the density given the latent variable $z$ as the input $\rho(z)$. To ensure values in $(0, 1)$, the estimated density is normalized using a sigmoid function $\bar{\rho}(z) = 1 / (1 + \exp(-(\rho(z) + \epsilon)/\tau ))$ with an appropriate offset $\epsilon$ and temperature $\tau$ parameters, and then used as the following to compute the output of the augmented policy  $\widetilde{\pi}$:
\begin{equation}
    \label{combination}
    \widetilde{\pi}(s) = \bar{\rho}(z)  \pi(s) + \left(1 - \bar{\rho}(z) \right)\pi_R(s).
\end{equation}

In the following sections, we first introduce our proposed equivariant mapping which maps raw visual observations into a latent space in which translation corresponds to the robot actions. Then, we introduce our method to estimate the density of the training observations in the latent space. Finally,  we describe how the recovery policy is constructed based on the learned equivariant mapping. 

\subsection{Learning an Equivariant Mapping}
We propose to learn a low-dimensional representation of the input visual observations by explicitly learning an equivariant mapping $E: S \rightarrow Z$. As we describe in section~\ref{sec:recovery_policy}, we exploit the equivariant property of the mapping to construct the recovery policy. Besides, learning a low-dimensional representation of visual inputs can also help in estimating the density of the training data. 

Given the transition of the MDP $T(s,a)$, we consider the following transition in the latent space: $T'(z,a) = z + a$. This is because the robot actions \emph{translate} the end-effector in the Euclidean space. We refer to Section \ref{future} for a discussion on how this can be generalized to actions beyond translations of an end-effector of a manipulator. We want to learn an equivariant mapping with respect to $T,T'$ i.e., $E$ has to satisfy the following version of Equation \ref{eq:equivariant_mapping}:
\begin{equation}
    \label{equiv}
    E(T(s,a)) = E(s) + a. 
\end{equation}
As shown in Figure~\ref{pipeline} (left), $E$ implements translational equivariance since it converts transitions into translations.
Note that Equation \ref{equiv} assumes that $A$ and $Z$ share the same ambient space $\mathbb{R}^n$. The equivariant model $E$ is trained on states and actions well-distributed within the environment i.e., a dataset $\mathcal{D}'$ is pre-collected independently from the specific task. As long as the scene composed of the extrinsic objects in the robot's environment remains constant, $E$ can be deployed in different tasks. In order to ensure a correct representation, $\mathcal{D}'$ needs to be distributed as uniformly as possible. The mapping $E$ is parameterized by a neural network $E = E_\varphi$ with output space $\mathbb{R}^n$ and optimized by minimizing the following objective function on the dataset $\mathcal{D}'$:

\begin{equation}
  \label{eqiloss}
  \varphi^* = \textnormal{argmin}_{\varphi} \sum_{(s,a, s') \in \mathcal{D}' } \| E_\varphi(s') - E_\varphi(s) -a \|^ 2.
\end{equation}

\subsection{Estimating the Density of the Training States }\label{densityest}
We estimate the probability density of the agent being within the support of the training data by learning a parametric \emph{density estimator}. The density estimator needs to be conditioned on the position of the manipulation objects for the task. It is thus conditioned on the image observation for the initial configuration of the manipulation task (Figure \ref{pipeline}, right). We use \emph{Mixture Density Networks} (MDN) \cite{bishop1994mixture}, which estimate a conditional Gaussian mixture density. The MDN outputs the density in the form of means $\mu_i$, (diagonal) co-variances $\sigma_i$ and weights $w_i$ of a mixture of Gaussians $\mathcal{N}(z; \  \mu_i, \sigma_i)$. The density in the point $z$ can then be computed as follows:
\begin{equation}
\label{mdn}
    \rho(z) = \sum_{i=1}^N w_i\mathcal{N}(z; \  \mu_i, \sigma_i).
\end{equation}
The MDN is trained by minimizing the average negative log-likelihood of $\rho$ over the observations in the training dataset $\mathcal{D}$.

\subsection{Recovery Policy}
\label{sec:recovery_policy}
The recovery policy is responsible to output actions that bring the agent closer to states within the support of the training data. 
This is done by finding an action that brings the agent into a state with higher estimated density. This is equivalent to performing gradient ascent on the estimated density function in the latent space. 
Because of the translational-equivariant property of our mapping, i.e., $z' = E(s') = E(s) + a = z + a$, the gradient $\nabla_z \rho(z)$ is equal to a robot action that moves the agent to higher density states. Therefore, we can simply define the output of the recovery policy for an observation $s$ as:
\begin{equation}
    \pi_R(s) = \eta \nabla_z \ \rho(E(s))
\end{equation}
where $\eta$ is a scale parameter. 
Once the density of states has been estimated in the latent space, the recovery policy can be implemented accordingly without any further training phase.

\section{Experiments}

In order to assess the effectiveness of the recovery of the proposed model, we present the results of a number of experiments. First, the experimental setup is described, then details on each of the experiments are presented.

\begin{itemize}
    \item The first experiment compares the performances of a BC agent with and without recovery on a robotic manipulation task.
    \item The second experiment tests the ability of a BC agent learned from noisy data, with and without recovery, in performing the same task.
    \item The third experiment compares the ability of BC, with and without recovery, in resuming a task if brought forcefully OOD.
    \item The fourth experiment involves solving a different task. The goal of this experiment is to show that the representation is agnostic to the task.
\end{itemize}

\subsection{Experimental Setup}

All the experiments are performed in the real world using a YuMi-IRB 14000 collaborative robot by ABB. We record all the data through teleoperation of the robot by a human. 
To implement the teleoperation system we used a virtual reality (VR) system connected to the robot's controllers. Teleoperation through VR has, in fact, proved to be a viable option for robotics applications thanks mainly to its ease in use and speed of data collection \cite{koganti2018virtual,mandlekar2021matters,zhang2018deep,whitney2018ros}. In particular, for our experiments, we interface an Oculus Quest 2 device to the robot's operating system.

To record the data, the human operator stands in front of the robot and operates the VR hand controller to command desired velocities to the end-effector of the robotic arm. Commands are thus mirrored with respect to the human perspective. Velocity commands are sent with a frequency of 10Hz to the robot that then translates them to the equivalent joint velocities. In all of the experiments, the velocities are just translations in space as no angular velocities are considered, meaning that $A \subseteq \mathbb{R}^3$. 
Images, representing the state of the system by a static camera placed in front of the robot in coordination with the VR commands.

The setup is the same across all experiments and its shown in Figure \ref{fig:trjpick}. The robot is placed in front of a table where two objects are placed, a small plastic orange cylinder, the \emph{manipulated object}, and a bigger plastic yellow cylinder, the \emph{target object}. Throughout all the experiments the target object is never moved while the robot needs to interact with the manipulated object.

\subsection{Networks' Architectures}

The equivariant encoder $E$ is parameterized by 5 convolutional layers with 64 channels except the last one with 8 followed by a hidden fully-connected layer with 64 units, every hidden layer is followed by a ReLU activation function. The network is trained with a learning rate of $10^{-3}$ using the Adam optimizer. Input images are cropped, resized and normalized before being fed to the network.

The policy is parameterized by a neural network with a ResNet18 backbone pre-trained on ImageNet and a randomly initialized fully connected head with 2 hidden layers of 64 units each. The MDN density estimator is parameterized by a CNN model of 4 layers with 64 channels and one with 8 channels followed by 3 output layers for the mean, diagonal variance and weights of the mixture of Gaussians. The MDN is trained, as stated in Section \ref{densityest}, to minimize the negative log-likelihood of the latent representation of the equivariant encoder. The MDN is conditioned on the initial image of each trajectory and the gripper state. The reason being that the model needs to be aware of the position of the object to be picked up to output a conditional density but it should not have access to the position of the gripper. There is, in fact, a functional dependency between images where the gripper is shown and the density estimate itself. By conditioning the MDN on the current image during the roll-out, the density would collapse. Conditioning the model on the gripper state is also needed as the position of the gripper should be considered in-distribution or not based on the object being grasped or not, respectively. Training an MDN is notoriously unstable \cite{makansi2019overcoming} and we found that adding a deconvolution decoder that maps a middle representation of the model into a reconstruction of the original image can stabilize the training. Both the policy and the MDN are trained using the Adam optimizer with a learning rate of $10^{-4}$. The other hyper-parameters used for this experiment are the following: $\eta$ = 0.05, $\epsilon$ = 2.0 and $\tau$ = 0.5.

\subsection{Equivariant Encoder}
The equivariant representation used for the density estimator is agnostic to the task and can be learned a priori. 

For the dataset $\mathcal{D}'$, we collect 6 trajectories of the robotic arm moving uniformly in the space and interacting with the manipulated object for a total of 1741 steps. Interaction with the objects is needed in order to make the learned encoder invariant to its position. Equivariance is defined with respect to the robot's actions only and the learned representation should not be sensitive to the object's position.

\begin{figure}[t]
\centering
\includegraphics[width=.95\columnwidth]{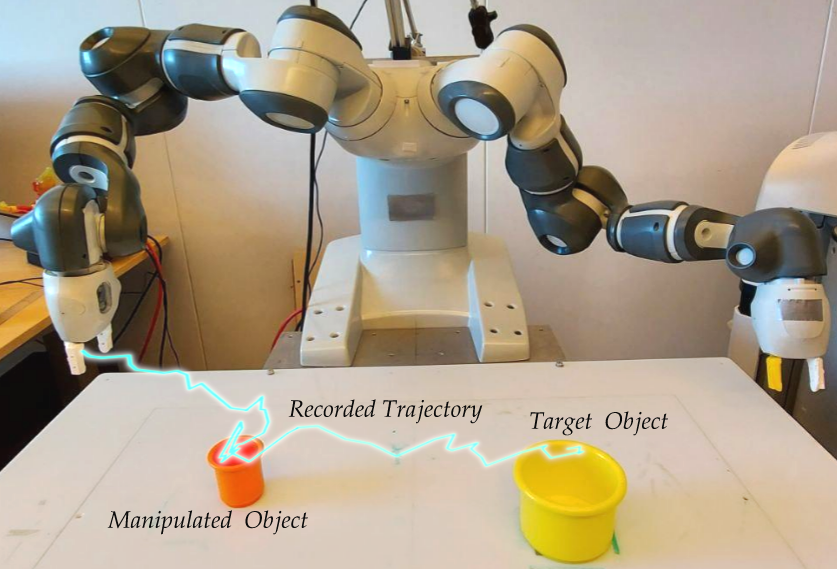}
\caption{Experimental setup for the pick-and-drop task. The small bin (\emph{manipulated object}) has to be dropped inside the larger one (\emph{target object}). A trajectory recorded by the expert via teleoperation is also displayed.}
\label{fig:trjpick}
\end{figure}

Because the actions are translations in the Euclidean 3D space the equivariant representation will correspond to points in 3D where the gripper of the robot is located (the central point of the actual movement). The representation preserves distances so the scale is one to one with the actions' magnitude. This can be seen in Figure \ref{fig:trjpick} where the expert's trajectories have been mapped into the latent space of the encoder and projected in 2D. In fact, the shape of the table and the relative position of the objects are maintained. For interpretability, we force the initial configuration of the robot to be the origin of the representation by including a second term in the loss of Equation \ref{eqiloss}. 

\subsection{Pick-and-Drop Experiment}

\subsubsection{Experiment description}
In the first experiment, the goal is for the robot to pick the manipulated object, move it on top of the target object and drop it inside. Here the actions are the combination of the gripper's velocity and a binary value representing the state of the gripper (either open or close). The initial position of the manipulated object is initialized randomly in the first half of the table while the target object is kept fixed on the other half of the table. A dataset of 120 trajectories of a human demonstrating the task is used to train the model. The dataset accounts for 5526 steps in total and is used to train both the behavioral cloning policy and the density estimator. 
The demonstrations are collected using the VR teleoperation system described above. However, these demonstrations cannot be assumed optimal due to the noise in the teleoperation system and the non-Markovianity of the environment. In fact, there are two elements here that break the Markovian assumption. The first is the current velocity and acceleration the robot has before giving it a command. The next state will vary depending on these properties that are not inferable from one single image. The second element is the inverse kinematic module of the robot. The resulting actual displacement of the end effector will also depend on the current state of the joints that is not fully observable from the images.

\subsubsection{Results}
In the first set of experiments, we test the proposed method against the imitation learning policy without the recovery term. The models are evaluated on 20 trials with the manipulated object in different positions. Performances are based on their ability to successfully grasp the manipulated object and their ability to then drop it inside the target bin. Table \ref{pick} shows the results of this experiment. The standard BC model manages to complete the task only one-fourth of the time. On the other hand, coupling the same policy with the proposed recovery lets the agent adjust its position every time it makes a mistake that would bring it OOD. This results in a much higher success rate.

Further, to assess the robustness of the recovery policy, we test it in noisy conditions. We train a second BC policy on the same dataset but with all the actions shifted by one step with respect to the images. We effectively simulate a data gathering scenario where there is a delay between the camera sensor and the actual movement of the robot. Two subsequent states are not connected by the saved action. However, because of the uniformity of the task, the real action does not differ considerably and the overall motion of the agent keeps a similar behavior. Nonetheless, a policy trained on this sub-optimal dataset does not manage to solve the task even once. On the other hand, by coupling the same policy with the proposed recovery module the performances are quite unchanged with respect to the correct dataset case, see Table \ref{pick}.

\begin{table}[t]
    \centering
    \setlength\extrarowheight{1.5pt}
    \begin{tabular}{ p{3cm} p{1.8cm} p{1.8cm}  }
    \toprule 
     \textsc{Model} & \textsc{Grasp} & \textsc{Drop} \\
     \hline
     \hline
     \multicolumn{3}{c}{\textsc{Pick-and-Drop}} \\
     \hline
     BC              & 25\% & 25\% \\
     BC with Recovery& 70\% & 55\% \\
     \hline
     \multicolumn{3}{c}{\textsc{Shifted Actions Pick-and-Drop}} \\
     \hline
     BC              &  0\% &  0\% \\
     BC with Recovery& 70\% & 50\% \\
     \hline
     \multicolumn{3}{c}{\textsc{Perturbed Pick-and-Drop}} \\
     \hline
     BC              &   30\% & 10\% \\
     BC with Recovery&  100\% & 70\% \\
    \bottomrule
    \end{tabular}
    \caption{Results of the pick-and-drop task for a standard behavioral cloning policy with and without the proposed recovery. The models are compared in their ability of picking the manipulated object correctly and dropping it inside the target object. The table shows results on models learned on correct demonstrations as well as demonstrations with actions that are shifted by one time step with respect to the corresponding images. Models are also tested on their ability to overcome perturbations while performing the task.}
    \label{pick}
\end{table}

Lastly, we test the BC policy with and without recovery by applying a random displacement to the robot. At the beginning of the task, we move the gripper in a random direction and let it continue the task from there. The new position could be outside of the training distribution, making the imitation learning policy behave randomly. The recovery policy instead can climb back to the training manifold and continue with the task normally. The object is initialized in a position where the imitation learning agent is able to complete the task. As shown in Table \ref{pick}, the learned policy suffers considerably from this kind of perturbations. On the other hand, when coupled with the proposed recovery it can always get back in and most of the time complete the task.

\subsection{Pushing Experiment}
The second set of experiments involves the same environment with unchanged dynamics. The goal is for the agent to insert the gripper's tip into the manipulated object and push it all the way towards the target object. In this experiment, the gripper is always closed. Because the robotic arm moves in the same way and only the manipulated object is moved throughout the roll-outs, the encoder can be used without retraining. Both the imitation learning policy and the density estimator have to be retrained on the new demonstrations. For this experiment, a new dataset of 60 demonstrations is collected for a total of 3895 steps. Results in Table \ref{push} show that the encoder can be used without retraining and that the recovery policy increases the performances of the learned policy.

\begin{table}[t]
    \centering
    \setlength\extrarowheight{1.5pt}
    \begin{tabular}{ p{3cm} p{1.8cm}   }
     \toprule
     \multicolumn{2}{c}{\textsc{Push}} \\
     \hline
     \hline
     \textsc{Model} & \textsc{Complete} \\
     \hline
     BC              &  20\% \\
     BC with Recovery&  25\% \\
     \bottomrule
    \end{tabular}
    \caption{Results of the pushing task for standard behavioral cloning policy with and without the proposed recovery. The models are compared in their ability of inserting the tip of the gripper within the manipulated object correctly and pushing it towards the target.}
    \label{push}
\end{table}

\iffalse
\begin{table}[t]
    \centering
    \begin{tabular}{ |p{3cm}||p{1.8cm}|p{1.8cm}|  }
     \hline
     \multicolumn{3}{|c|}{Push Task} \\
     \hline
     Model&Insertion &Complete\\
     \hline
     BC              &  60\% & 20\% \\
     BC with Recovery&  45\% & 25\% \\
     \hline
    \end{tabular}
    \caption{Results of the pushing task for standard behavioral cloning policy with and without the proposed recovery. The models are compared in their ability of inserting the tip of the gripper within the manipulated object correctly and pushing it towards the target.}
    \label{push}
\end{table}
\fi
\section{Conclusions and Future Work}\label{future}

We proposed to couple an agent learned on experts' demonstrations with a recovery policy to keep it within the training data. This is achieved by explicitly modeling the training distribution with a density estimator and bypassing the agent's action whenever the current state is detected to be OOD. By training an encoder to be equivariant to the agent's actions, the recovery policy can be formulated as a form of gradient ascent on the density estimate.

We applied the proposed methodology to a robotic manipulator whose actions correspond to Euclidean translations. As a possible extension, more complex actions could be considered such as rotations of joins and end-effectors. This would involve \emph{Lie groups} beyond the Euclidean space such as the group of rotations $\textnormal{SO}(n)$, $n=2,3$. A further extension of the framework lies in designing more complex recovery strategies than pure gradient ascent in order to smooth the resulting movement.

\section{Acknowledgements}

This work has been supported by the Swedish Research Council, Knut and Alice Wallenberg Foundation, European Research Council (BIRD-884807) and H2020 CANOPIES.

%%%%%%%%%%%%%%%%%%%%%%%%%%%%%%%%%%%%%%%%%%%%%%%%%%%%%%%%%%%%%%%%%%%%%%%%%%%%%%%%

\bibliography{references}

\begin{thebibliography}{10}
\providecommand{\url}[1]{#1}
\csname url@rmstyle\endcsname
\providecommand{\newblock}{\relax}
\providecommand{\bibinfo}[2]{#2}
\providecommand\BIBentrySTDinterwordspacing{\spaceskip=0pt\relax}
\providecommand\BIBentryALTinterwordstretchfactor{4}
\providecommand\BIBentryALTinterwordspacing{\spaceskip=\fontdimen2\font plus
\BIBentryALTinterwordstretchfactor\fontdimen3\font minus
  \fontdimen4\font\relax}
\providecommand\BIBforeignlanguage[2]{{%
\expandafter\ifx\csname l@#1\endcsname\relax
\typeout{** WARNING: IEEEtran.bst: No hyphenation pattern has been}%
\typeout{** loaded for the language `#1'. Using the pattern for}%
\typeout{** the default language instead.}%
\else
\language=\csname l@#1\endcsname
\fi
#2}}

\bibitem{levine2018learning}
S.~Levine, P.~Pastor, A.~Krizhevsky, J.~Ibarz, and D.~Quillen, ``Learning
  hand-eye coordination for robotic grasping with deep learning and large-scale
  data collection,'' \emph{The International journal of robotics research},
  vol.~37, no. 4-5, pp. 421--436, 2018.

\bibitem{levine2020offline}
S.~Levine, A.~Kumar, G.~Tucker, and J.~Fu, ``Offline reinforcement learning:
  Tutorial, review, and perspectives on open problems,'' \emph{arXiv preprint
  arXiv:2005.01643}, 2020.

\bibitem{hussein2017imitation}
A.~Hussein, M.~M. Gaber, E.~Elyan, and C.~Jayne, ``Imitation learning: A survey
  of learning methods,'' \emph{ACM Computing Surveys (CSUR)}, vol.~50, no.~2,
  pp. 1--35, 2017.

\bibitem{dagger}
S.~Ross, G.~Gordon, and D.~Bagnell, ``A reduction of imitation learning and
  structured prediction to no-regret online learning,'' in \emph{Proceedings of
  the fourteenth international conference on artificial intelligence and
  statistics}.\hskip 1em plus 0.5em minus 0.4em\relax JMLR Workshop and
  Conference Proceedings, 2011, pp. 627--635.

\bibitem{zhou2020plas}
W.~Zhou, S.~Bajracharya, and D.~Held, ``Plas: Latent action space for offline
  reinforcement learning,'' \emph{arXiv preprint arXiv:2011.07213}, 2020.

\bibitem{kumar2020conservative}
A.~Kumar, A.~Zhou, G.~Tucker, and S.~Levine, ``Conservative q-learning for
  offline reinforcement learning,'' \emph{Advances in Neural Information
  Processing Systems}, vol.~33, pp. 1179--1191, 2020.

\bibitem{yu2021combo}
T.~Yu, A.~Kumar, R.~Rafailov, A.~Rajeswaran, S.~Levine, and C.~Finn, ``Combo:
  Conservative offline model-based policy optimization,'' \emph{arXiv preprint
  arXiv:2102.08363}, 2021.

\bibitem{ghadirzadeh2016sensorimotor}
A.~Ghadirzadeh, J.~B{\"u}tepage, A.~Maki, D.~Kragic, and M.~Bj{\"o}rkman, ``A
  sensorimotor reinforcement learning framework for physical human-robot
  interaction,'' in \emph{2016 IEEE/RSJ International Conference on Intelligent
  Robots and Systems (IROS)}.\hskip 1em plus 0.5em minus 0.4em\relax IEEE,
  2016, pp. 2682--2688.

\bibitem{kahn2017uncertainty}
G.~Kahn, A.~Villaflor, V.~Pong, P.~Abbeel, and S.~Levine, ``Uncertainty-aware
  reinforcement learning for collision avoidance,'' \emph{arXiv preprint
  arXiv:1702.01182}, 2017.

\bibitem{survey1}
T.~Osa, J.~Pajarinen, G.~Neumann, J.~A. Bagnell, P.~Abbeel, and J.~Peters, ``An
  algorithmic perspective on imitation learning,'' \emph{arXiv preprint
  arXiv:1811.06711}, 2018.

\bibitem{pmlr-v9-ross10a}
S.~Ross and D.~Bagnell, ``Efficient reductions for imitation learning,'' in
  \emph{Proceedings of the Thirteenth International Conference on Artificial
  Intelligence and Statistics}, 2010, pp. 661--668.

\bibitem{abbeel2010inverse}
P.~Abbeel and A.~Y. Ng, ``Inverse reinforcement learning.'' 2010.

\bibitem{ho2016generative}
J.~Ho and S.~Ermon, ``Generative adversarial imitation learning,''
  \emph{Advances in neural information processing systems}, vol.~29, 2016.

\bibitem{precup2000eligibility}
D.~Precup, ``Eligibility traces for off-policy policy evaluation,''
  \emph{Computer Science Department Faculty Publication Series}, p.~80, 2000.

\bibitem{kumar2019stabilizing}
A.~Kumar, J.~Fu, M.~Soh, G.~Tucker, and S.~Levine, ``Stabilizing off-policy
  q-learning via bootstrapping error reduction,'' \emph{Advances in Neural
  Information Processing Systems}, vol.~32, 2019.

\bibitem{fujimoto2019off}
S.~Fujimoto, D.~Meger, and D.~Precup, ``Off-policy deep reinforcement learning
  without exploration,'' in \emph{International Conference on Machine
  Learning}.\hskip 1em plus 0.5em minus 0.4em\relax PMLR, 2019, pp. 2052--2062.

\bibitem{chen2022latent}
X.~Chen, A.~Ghadirzadeh, T.~Yu, Y.~Gao, J.~Wang, W.~Li, B.~Liang, C.~Finn, and
  C.~Zhang, ``Latent-variable advantage-weighted policy optimization for
  offline rl,'' \emph{arXiv preprint arXiv:2203.08949}, 2022.

\bibitem{yu2020mopo}
T.~Yu, G.~Thomas, L.~Yu, S.~Ermon, J.~Y. Zou, S.~Levine, C.~Finn, and T.~Ma,
  ``Mopo: Model-based offline policy optimization,'' \emph{Advances in Neural
  Information Processing Systems}, vol.~33, pp. 14\,129--14\,142, 2020.

\bibitem{brunke2021safe}
L.~Brunke, M.~Greeff, A.~W. Hall, Z.~Yuan, S.~Zhou, J.~Panerati, and A.~P.
  Schoellig, ``Safe learning in robotics: From learning-based control to safe
  reinforcement learning,'' \emph{Annual Review of Control, Robotics, and
  Autonomous Systems}, vol.~5, 2021.

\bibitem{lee2020guided}
M.~A. Lee, C.~Florensa, J.~Tremblay, N.~Ratliff, A.~Garg, F.~Ramos, and D.~Fox,
  ``Guided uncertainty-aware policy optimization: Combining learning and
  model-based strategies for sample-efficient policy learning,'' in \emph{2020
  IEEE International Conference on Robotics and Automation (ICRA)}.\hskip 1em
  plus 0.5em minus 0.4em\relax IEEE, 2020, pp. 7505--7512.

\bibitem{thananjeyan2021recovery}
B.~Thananjeyan, A.~Balakrishna, S.~Nair, M.~Luo, K.~Srinivasan, M.~Hwang, J.~E.
  Gonzalez, J.~Ibarz, C.~Finn, and K.~Goldberg, ``Recovery rl: Safe
  reinforcement learning with learned recovery zones,'' \emph{IEEE Robotics and
  Automation Letters}, vol.~6, no.~3, pp. 4915--4922, 2021.

\bibitem{srinivasan2020learning}
K.~Srinivasan, B.~Eysenbach, S.~Ha, J.~Tan, and C.~Finn, ``Learning to be safe:
  Deep rl with a safety critic,'' \emph{arXiv preprint arXiv:2010.14603}, 2020.

\bibitem{thananjeyan2020safety}
B.~Thananjeyan, A.~Balakrishna, U.~Rosolia, F.~Li, R.~McAllister, J.~E.
  Gonzalez, S.~Levine, F.~Borrelli, and K.~Goldberg, ``Safety augmented value
  estimation from demonstrations (saved): Safe deep model-based rl for sparse
  cost robotic tasks,'' \emph{IEEE Robotics and Automation Letters}, vol.~5,
  no.~2, pp. 3612--3619, 2020.

\bibitem{wilcox2022ls3}
A.~Wilcox, A.~Balakrishna, B.~Thananjeyan, J.~E. Gonzalez, and K.~Goldberg,
  ``Ls3: Latent space safe sets for long-horizon visuomotor control of sparse
  reward iterative tasks,'' in \emph{Conference on Robot Learning}.\hskip 1em
  plus 0.5em minus 0.4em\relax PMLR, 2022, pp. 959--969.

\bibitem{mitsioni2021safe}
I.~Mitsioni, P.~Tajvar, D.~Kragic, J.~Tumova, and C.~Pek, ``Safe data-driven
  contact-rich manipulation,'' in \emph{2020 IEEE-RAS 20th International
  Conference on Humanoid Robots (Humanoids)}.\hskip 1em plus 0.5em minus
  0.4em\relax IEEE, 2021, pp. 120--127.

\bibitem{li2018learning}
M.~Li, K.~Tahara, and A.~Billard, ``Learning task manifolds for constrained
  object manipulation,'' \emph{Autonomous Robots}, vol.~42, no.~1, pp.
  159--174, 2018.

\bibitem{lesort2018state}
T.~Lesort, N.~D{\'\i}az-Rodr{\'\i}guez, J.-F. Goudou, and D.~Filliat, ``State
  representation learning for control: An overview,'' \emph{Neural Networks},
  vol. 108, pp. 379--392, 2018.

\bibitem{chen2020adversarial}
X.~Chen, A.~Ghadirzadeh, M.~Bj{\"o}rkman, and P.~Jensfelt, ``Adversarial
  feature training for generalizable robotic visuomotor control,'' in
  \emph{2020 IEEE International Conference on Robotics and Automation
  (ICRA)}.\hskip 1em plus 0.5em minus 0.4em\relax IEEE, 2020, pp. 1142--1148.

\bibitem{hamalainen2019affordance}
A.~H{\"a}m{\"a}l{\"a}inen, K.~Arndt, A.~Ghadirzadeh, and V.~Kyrki, ``Affordance
  learning for end-to-end visuomotor robot control,'' in \emph{2019 IEEE/RSJ
  International Conference on Intelligent Robots and Systems (IROS)}.\hskip 1em
  plus 0.5em minus 0.4em\relax IEEE, 2019, pp. 1781--1788.

\bibitem{finn2016deep}
C.~Finn, X.~Y. Tan, Y.~Duan, T.~Darrell, S.~Levine, and P.~Abbeel, ``Deep
  spatial autoencoders for visuomotor learning,'' in \emph{2016 IEEE
  International Conference on Robotics and Automation (ICRA)}.\hskip 1em plus
  0.5em minus 0.4em\relax IEEE, 2016, pp. 512--519.

\bibitem{ghadirzadeh2017deep}
A.~Ghadirzadeh, A.~Maki, D.~Kragic, and M.~Bj{\"o}rkman, ``Deep predictive
  policy training using reinforcement learning,'' in \emph{2017 IEEE/RSJ
  International Conference on Intelligent Robots and Systems (IROS)}.\hskip 1em
  plus 0.5em minus 0.4em\relax IEEE, 2017, pp. 2351--2358.

\bibitem{assael2015data}
J.-A.~M. Assael, N.~Wahlstr{\"o}m, T.~B. Sch{\"o}n, and M.~P. Deisenroth,
  ``Data-efficient learning of feedback policies from image pixels using deep
  dynamical models,'' \emph{arXiv preprint arXiv:1510.02173}, 2015.

\bibitem{ha2018world}
D.~Ha and J.~Schmidhuber, ``World models,'' \emph{arXiv preprint
  arXiv:1803.10122}, 2018.

\bibitem{goroshin2015learning}
R.~Goroshin, M.~F. Mathieu, and Y.~LeCun, ``Learning to linearize under
  uncertainty,'' \emph{Advances in neural information processing systems},
  vol.~28, 2015.

\bibitem{watter2015embed}
M.~Watter, J.~Springenberg, J.~Boedecker, and M.~Riedmiller, ``Embed to
  control: A locally linear latent dynamics model for control from raw
  images,'' \emph{Advances in neural information processing systems}, vol.~28,
  2015.

\bibitem{zhang2019solar}
M.~Zhang, S.~Vikram, L.~Smith, P.~Abbeel, M.~Johnson, and S.~Levine, ``Solar:
  Deep structured representations for model-based reinforcement learning,'' in
  \emph{International Conference on Machine Learning}.\hskip 1em plus 0.5em
  minus 0.4em\relax PMLR, 2019, pp. 7444--7453.

\bibitem{jaques2021newtonianvae}
M.~Jaques, M.~Burke, and T.~M. Hospedales, ``Newtonianvae: Proportional control
  and goal identification from pixels via physical latent spaces,'' in
  \emph{Proceedings of the IEEE/CVF Conference on Computer Vision and Pattern
  Recognition}, 2021, pp. 4454--4463.

\bibitem{kipf2019contrastive}
T.~Kipf, E.~van~der Pol, and M.~Welling, ``Contrastive learning of structured
  world models,'' \emph{arXiv preprint arXiv:1911.12247}, 2019.

\bibitem{cohen2016group}
T.~Cohen and M.~Welling, ``Group equivariant convolutional networks,'' in
  \emph{International conference on machine learning}.\hskip 1em plus 0.5em
  minus 0.4em\relax PMLR, 2016, pp. 2990--2999.

\bibitem{marchetti2022equivariant}
G.~L. Marchetti, G.~Tegn{\'e}r, A.~Varava, and D.~Kragic, ``Equivariant
  representation learning via class-pose decomposition,'' \emph{arXiv preprint
  arXiv:2207.03116}, 2022.

\bibitem{bishop1994mixture}
C.~M. Bishop, ``Mixture density networks,'' 1994.

\bibitem{koganti2018virtual}
N.~Koganti, A.~Rahman~HAG, Y.~Iwasawa, K.~Nakayama, and Y.~Matsuo, ``Virtual
  reality as a user-friendly interface for learning from demonstrations,'' in
  \emph{Extended Abstracts of the 2018 CHI Conference on Human Factors in
  Computing Systems}, 2018, pp. 1--4.

\bibitem{mandlekar2021matters}
A.~Mandlekar, D.~Xu, J.~Wong, S.~Nasiriany, C.~Wang, R.~Kulkarni, L.~Fei-Fei,
  S.~Savarese, Y.~Zhu, and R.~Mart{\'\i}n-Mart{\'\i}n, ``What matters in
  learning from offline human demonstrations for robot manipulation,''
  \emph{arXiv preprint arXiv:2108.03298}, 2021.

\bibitem{zhang2018deep}
T.~Zhang, Z.~McCarthy, O.~Jow, D.~Lee, X.~Chen, K.~Goldberg, and P.~Abbeel,
  ``Deep imitation learning for complex manipulation tasks from virtual reality
  teleoperation,'' in \emph{2018 IEEE International Conference on Robotics and
  Automation (ICRA)}.\hskip 1em plus 0.5em minus 0.4em\relax IEEE, 2018, pp.
  5628--5635.

\bibitem{whitney2018ros}
D.~Whitney, E.~Rosen, D.~Ullman, E.~Phillips, and S.~Tellex, ``Ros reality: A
  virtual reality framework using consumer-grade hardware for ros-enabled
  robots,'' in \emph{2018 IEEE/RSJ International Conference on Intelligent
  Robots and Systems (IROS)}.\hskip 1em plus 0.5em minus 0.4em\relax IEEE,
  2018, pp. 1--9.

\bibitem{makansi2019overcoming}
O.~Makansi, E.~Ilg, O.~Cicek, and T.~Brox, ``Overcoming limitations of mixture
  density networks: A sampling and fitting framework for multimodal future
  prediction,'' in \emph{Proceedings of the IEEE/CVF Conference on Computer
  Vision and Pattern Recognition}, 2019, pp. 7144--7153.

\end{thebibliography}
\bibliographystyle{IEEEtran}

\end{document}